\documentclass[11pt,oneside,a4paper]{article}
\usepackage{fullpage}
\usepackage[affil-it]{authblk}
\usepackage{amsmath}
\usepackage{amsthm}
\usepackage{amsfonts}
\usepackage{amssymb}
\usepackage{mathtools}
\usepackage{multirow}
\usepackage{url}
\usepackage{booktabs}
\usepackage{subfigure}
\usepackage{array}

\newcommand{\eg}{e.\,g., }
\newcommand{\ie}{i.\,e., }

\newcolumntype{H}{>{\setbox0=\hbox\bgroup}c<{\egroup}@{}}


\title{\textbf{Neural Graph Generator: Feature-Conditioned Graph Generation using Latent Diffusion Models}}

\author{Iakovos Evdaimon\textsuperscript{\rm 1},
Giannis Nikolentzos\textsuperscript{\rm 4}, Christos Xypolopoulos\textsuperscript{\rm 1,3}, Ahmed Kammoun \textsuperscript{\rm 1}, Michail Chatzianastasis\textsuperscript{\rm 1}, Hadi Abdine\textsuperscript{\rm 1}, and Michalis Vazirgiannis\textsuperscript{\rm 1,2}\\

\textnormal{\textsuperscript{\rm 1}\'Ecole Polytechnique, IP Paris, France}\\
\textnormal{\textsuperscript{\rm 2} Mohamed bin Zayed University of Artificial Intelligence, United Arab Emirates}\\
\textnormal{\textsuperscript{\rm 3} National Technical University of Athens, Greece}
\textnormal{\textsuperscript{\rm 4}University of Peloponnese, Greece}\\
}
\date{}

\begin{document}
\maketitle 

\begin{abstract}
Graph generation has emerged as a crucial task in machine learning, with significant challenges in generating graphs that accurately reflect specific properties. 
Existing methods often fall short in efficiently addressing this need as they struggle with the high-dimensional complexity and varied nature of graph properties.
In this paper, we introduce the \textbf{Neural Graph Generator}~(NGG), a novel approach which utilizes conditioned latent diffusion models for graph generation. 
NGG demonstrates a remarkable capacity to model complex graph patterns, offering control over the graph generation process. NGG employs a variational graph autoencoder for graph compression and a diffusion process in the latent vector space, guided by vectors summarizing graph statistics.
We demonstrate NGG's versatility across various graph generation tasks, showing its capability to capture desired graph properties and generalize to unseen graphs. 
We also compare our generator to the graph generation capabilities of different LLMs. 
This work signifies a shift in graph generation methodologies, offering a more practical and efficient solution for generating diverse graphs with specific characteristics.
\end{abstract}

\section{Introduction}
In recent years, the field of machine learning on graphs has witnessed an extensive growth, mainly due to the availability of large amounts of data represented as graphs.
Indeed, graphs arise naturally in several application domains such as in social networks, in chemo-informatics and in bio-informatics.
One of the most challenging tasks of machine learning on graphs is that of graph generation~\cite{zhu2022survey}.
Graph generation has attracted a lot of attention recently and its main objective is to create novel and realistic graphs.
For instance, in chemo-informatics, graph generative models are employed to generate novel, realistic molecular graphs which also exhibit desired properties (\eg high drug-likeness)~\cite{jin2018junction,zang2020moflow}.

Recently, there is a surge of interest in developing new graph generative models, and most of the proposed models typically fall into one of the following five families of models: (1) Auto-Regressive models; (2) Variational Autoencoders; (3) Generative Adversarial Networks; (4) Normalizing Flows; and (5) Diffusion models.
These models can capture the complex structural and semantic information of graphs, but focus mainly on specific types of graphs such as molecules~\cite{pmlr-v162-hoogeboom22a}, proteins~\cite{ingraham2019generative}, computer programs~\cite{brockschmidt2018generative} or patient trajectories~\cite{nikolentzos2023synthetic}.
Traditionally, in different application domains, there is a need for generating graphs that exhibit specific properties (\eg degree distribution, node triangle participation, community structure, etc.).
The different models that are commonly employed to generate these graphs, such as the Erd{\H{o}}s--R{\'e}nyi model and the Barab{\'a}si-Albert model, capture one or more network properties but neglect others. 
For example, the Barab{\'a}si-Albert model can generate graphs with a specific degree distribution, but ignores the rest of the graph's properties.

Recent graph generative models have improved in accurately representing various properties of real-world graphs. Nonetheless, to produce graphs adhering to specific properties, it is necessary first to compile a dataset of graphs that display these characteristics before proceeding to train the model using this dataset.
Thus, to generate different types of graphs, different models need to be trained, which is impractical.
This work aims to fill this gap.
Specifically, we develop a neural network model, so-called Neural Graph Generator (NGG), to generate graphs that exhibit specific properties.
We capitalize on recent advances in latent diffusion models to perform vector-conditioned graph generation.
The model applies the diffusion process not to the graph data but instead to an encoded latent representation of the graph.
We produce vectors that contain a summary of the statistics of each graph (\eg number of nodes, number of edges, clustering coefficient, etc.).
Those vectors guide the denoising part of the diffusion process.
We first pre-train a variational graph autoencoder which we use to map graphs into vectors and map vectors into graphs.
The diffusion is performed in the space of vectors.
To generate new graphs that exhibit specific properties, we sample a vector from the standard normal distribution, and we denoise this vector, while denoising is guided by the vector of graph properties.
Once the noise has been removed from the vector, we feed the emerging vector into the decoder of the pre-trained variational graph autoencoder to produce a graph.
The proposed model is extensively evaluated on two axes: i. tasks of graph generation conditioned on multiple graph properties ii. comparison to graph generation results of different state-of-the-art LLMs (Llama2, Llama3, Gemma, Mistral and GPT3.5).
Our results indicate that the NGG pretrained model can indeed generate graphs whose properties are similar to the ones that are given as input to the model and also it out performs LLMs in all aspects including error, accuracy and execution times where our approach is orders of magnitude faster than LLMs while it consumes much less memory in GPU. 

To summarize, our work makes the following contributions:
\begin{itemize}
    \item[(i)]  We introduce \textbf{Neural Graph Generator}, a novel graph generative model which leverages latent diffusion  for conditional graph generation. This model represents a significant shift from traditional graph generation methods, focusing on prompting with a vector that includes a set of diverse properties of the graph.
    \item[(ii)] We introduce a large-scale dataset of synthetic graphs that covers several different types of graphs on which our model was trained.
    This dataset can be used for pre-training any graph generative model in the future. 
    \item[(iii)] We extensively evaluate our model across various graph generation tasks, demonstrating its effectiveness in capturing specific graph properties, generalizing to larger graphs, and generating graphs from subsets of properties. 
    \item [(iv)]we perform - as part of the experimental evaluation of NGG - the first comprehensive evaluation of LLMs in graph generation tasks, covering many generic properties of graphs.
    
    \item[(v)]We release the pre-trained autoencoder, the pre-trained latent diffusion model, and the synthetic dataset of 1M graphs which are likely to be useful for both practitioners and the scientific community \footnote{\url{https://github.com/iakovosevdaimon/Graph-Generator/}}.
\end{itemize}


\section{Related Work}\label{sec:related_work}

\paragraph{Graph Generative Models.}
For the problem of graph generation, as discussed above, most models belong to one of the following five families: (1) Auto-regressive models; (2) Variational autoencoders; (3) Generative adversarial networks; (4) Normalizing flows; and (5) Diffusion models.
Auto-regressive models assume a specific node ordering and generate graphs in a sequential fashion.
GraphRNN~\cite{you2018graphrnn}, GraphGen~\cite{goyal2020graphgen} and GraphGen-Redux~\cite{bacciu2021graphgen} are three examples of auto-regressive models.
GraphRNN generates nodes and its associated edges sequentially, while the other two models generate edges and their endpoint nodes sequentially.
Variational autoencoders consist of two modules.
First, the encoder which maps the input data to a space that corresponds to the parameters of a Gaussian distribution.
Typically, this distribution is encouraged to be similar to the standard normal distribution.
Second, the decoder which maps from the latent space to the input space.
In the graph domain, GraphVAE~\cite{simonovsky2018graphvae} is a model whose encoder is an instance of a message passing graph neural network, while its decoder is a simple multi-layer perceptron.
There are models that embed graphs into distributions different from Gaussian such as DGVAE~\cite{li2020dirichlet} which utilizes the Dirichlet distributions as priors on the latent variables and the latent variables represent graph cluster memberships.
Variational autoencoders have also achieved success in the field of chemoinformatics~\cite{jin2018junction,samanta2020nevae}.
Similar to variational autoencoders, generative adversarial networks also consist of two main components, a generator whose objective is to generate realistic samples and a discriminator whose objective is to distinguish real samples from synthetic ones.
Most models from this family typically consist of a generator that takes vectors sampled from a standard normal distribution and processes them with a multi-layer perceptron to generate a graph, and of a discriminator which is an instance of a message passing graph neural network~\cite{de2018molgan,polsterl2021adversarial}.
Architectures based on normalizing flows explicitly model a probability distribution by leveraging a method that uses the change-of-variable law of probabilities to transform a simple distribution into a complex one by a sequence of invertible and differentiable mappings. 
Normalizing flows can be applied to graphs by designing message passing mechanisms that are exactly reversible~\cite{liu2019graph}.
Such models have been applied to molecular data.
For example, MoFlow~\cite{zang2020moflow} generates chemical bonds through a Glow-based model, and atoms given bonds using a graph conditional flow.
Diffusion models gradually add noise to the input data in the forward diffusion process, and then learn to remove the noise in the reverse diffusion process.
There exist models that add Gaussian noise to the graph's adjacency matrix and binarize the continuous values of the output of the reverse diffusion process to produce valid graphs~\cite{niu2020permutation,jo2022score}.
More recent models apply a discrete diffusion process that progressively adds noise by adding or removing edges and changing the nodes' and/or edges' types~\cite{vignac2022digress}.
Recently, several studies have focused on the potential of large language models (LLMs) for graph generation, demonstrating promising preliminary capabilities in both rule-based and distribution-based graph generation~\cite{yao2024exploringpotentiallargelanguage}.

\paragraph{Conditional Graph Generation.}
One of the main objectives of graph generative models is conditional generation, which is the process of generating a graph that satisfies a specific label or property.
Typically, a conditional code is introduced that explicitly controls the property of generated graphs.
Such models are popular in the field of chemo-informatics where novel molecules that
possess desired chemical properties (\eg high binding affinity against a target protein) need to be generated.
For variational autoencoders, a simple approach is to feed the concatenation of the conditional code and the latent graph representation to the decoder~\cite{simonovsky2018graphvae}.
The conditional code can also be concatenated to node representations sampled from a single distribution~\cite{yang2019conditional}.
In models that generate new nodes and edges sequentially, once a new node is created, its feature vector must be initialized, and the conditional code could be concatenated to that feature vector~\cite{li2018learning}.
There exist models whose message passing scheme is modified to include the conditional code~\cite{li2018multi}.
MOOD~\cite{lee2023exploring} utilizes the gradient of a property prediction network to guide the sampling process to domains that are likely to satisfy specific properties.
There are also models that employ regularization schemes to learn disentangled representations where each dimension focuses on a specific property~\cite{du2022interpretable}.

\paragraph{Latent Diffusion Models.}
Diffusion models are generative models that have recently gained significant attention~\cite{ho2020denoising,sohl2015deep,vincent2011connection}.
Diffusion models have achieved state-of-the-art performance in several tasks such as in conditional image generation~\cite{dhariwal2021diffusion,song2020score}, image colorization~\cite{saharia2022palette} and image super resolution~\cite{saharia2022image}, just to name a few.
Latent diffusion models were introduced to reduce the computational complexity of diffusion models.
These models use some autoencoding component to project input data (\eg images) into some latent space and perform the diffusion process in that space~\cite{rombach2022high}.
Latent diffusion models can thus be trained on limited computational resources while retaining the quality of diffusion models.
Latent diffusion models have been applied to different problems such as image synthesis~\cite{rombach2022high}, video synthesis~\cite{blattmann2023align} and image reconstruction~\cite{takagi2023high}.
Recently, latent diffusion models have been introduced in the molecular geometry domain, and have been evaluated in the task of 3D molecule generation~\cite{xu2023geometric}, but also for generating protein backbone structures~\cite{fu2023latent}.
These two works are the closest to our method.
Even though architecture-wise our model is similar to the models presented in these papers, we follow an entirely different research direction and focus on conditional generation of general graphs (similar to traditional graph generators) instead of certain classes of graphs such as molecules and proteins. 



\section{Neural Graph Generator}\label{sec:methodology}
\begin{figure*}[t]
    \begin{center}
    \centerline{\includegraphics[width=.9\linewidth]{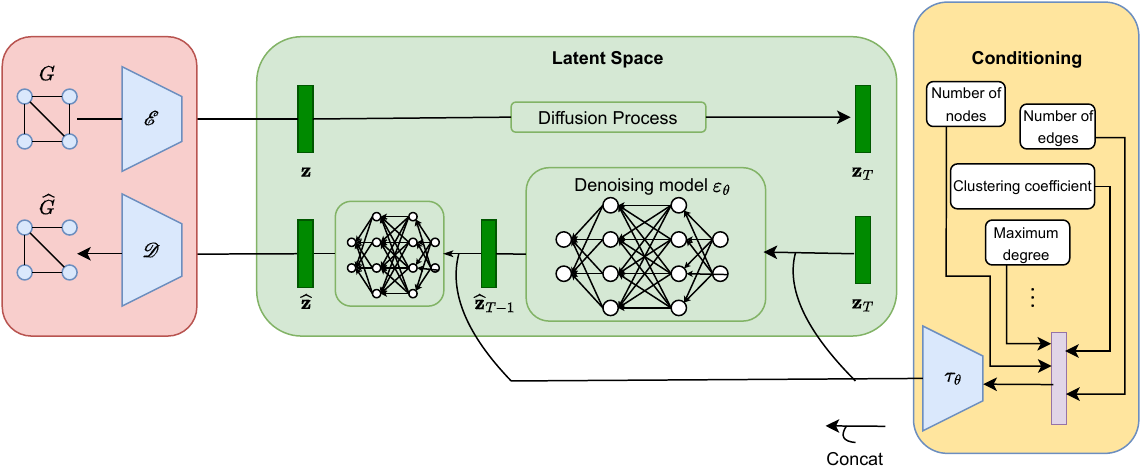}}
    \caption{Overview of the proposed architecture. The variational graph autoencoder is responsible for generating a compressed latent representation $\boldsymbol{z}$ for each graph $\boldsymbol{G}$. Those representations are fed to the diffusion model which operates in that latent space adding noise to $\boldsymbol{z}$ resulting to $\boldsymbol{z}_T$ . The denoising process is conditioned on the encoding (output of $\tau_\theta$) of the vector that contains the graph's properties. The output of the diffusion model is passed on to the decoder which generates a graph.}
    \label{fig:model}
    \end{center}
    \vspace{-.2cm}
\end{figure*}

In this section, we first introduce some key notation for graphs, and we then present the two main components of the proposed NGG model: ($1$) the variational graph autoencoder which produces a compressed latent representation for each graph; and ($2$) the diffusion model which performs diffusion in the latent space and which can be conditioned on various inputs (vector of graph properties in our case).
An overview of the proposed model is given in Figure~\ref{fig:model}.

\subsection{Notation}
Let $[n] = \{1,\ldots,n\} \subset \mathbb{N}$ for $n \geq 1$.
Let $G = (V,E)$ be an undirected graph, where $V$ is the set of nodes and $E$ is the edge set.
We will denote by $n$ the number of vertices and by $m$ the number of edges, \ie $n = |V|$ and $m = |E|$.
The neighborhood $\mathcal{N}(v)$ of a node $v$ is the set of all vertices adjacent to $v$.
Hence, $\mathcal{N}(v) = \{ u \, | \, (v,u) \in E\}$ where $(v,u)$ is an edge between the vertices $v$ and $u$ of $V$.
The adjacency matrix $\mathbf{A} \in \mathbb{R}^{n \times n}$ of a graph $G$ is a symmetric (typically sparse) matrix that is used to encode edge information in the graph.
The element of the $i^{th}$ row and $j^{th}$ column is equal to the weight of the edge between the vertices $v_i$ and $v_j$ if such an edge exists, and $0$ otherwise.
The degree $d(v)$ of a node $v$ is equal to the number of edges that are adjacent to the node, \ie $d(v) = |\mathcal{N}(v)|$.
In different applications, nodes of graphs are annotated with feature vectors.
We use $\mathbf{X} \in \mathbb{R}^{n \times d}$ to denote the node features where $d$ is the feature dimension size.
The feature of a given node $v_i$ corresponds to the $i^{th}$ row of $\mathbf{X}$.

\subsection{Graph Compression}
To map graphs into low-dimensional representations, we capitalize on previous work, and we use an autoencoder~\cite{simonovsky2018graphvae}.
More precisely, given a graph $G$, the encoder $\mathcal{E}$ encodes the graph into a latent representation $\mathbf{z} = \mathcal{E}(G)$, and the decoder $\mathcal{D}$ reconstructs the graph from the latent representation, giving $\hat{G} = \mathcal{D}(\mathbf{z}) = \mathcal{D}(\mathcal{E}(G))$, where
$\mathbf{z} \in \mathbb{R}^d$.

The encoder $\mathcal{E}$ corresponds to a message passing neural network which consists of GIN layers~\cite{xu2019how}.
Node representations are updated as follows:
\begin{equation*}
    \mathbf{h}_v^{(k)} = \text{MLP}^{(k)} \Bigg( \Big( 1 + \epsilon^{(k)} \Big) \mathbf{h}_v^{(k-1)} + \sum_{u \in \mathcal{N}(v)} \mathbf{h}_u^{(k-1)} \Bigg)
\end{equation*}
where $k$ denotes the layer and $\epsilon$ is a trainable parameter.
Note that the nodes are initially annotated with features extracted from the eigenvectors of the normalized Laplacian matrix.
These features are stored in matrix $\mathbf{X}$, while $\mathbf{h}_v^{(0)}$ is equal to the row of matrix $\mathbf{X}$ that corresponds to node $v$.
To produce a representation for the entire graph, we use the sum operator as follows: 
\begin{equation*}
    \mathbf{h}_G = \sum_{v \in V} \mathbf{h}_v^{(K)}
\end{equation*}    
where $K$ is the number of layers.
To avoid arbitrarily high-variance latent spaces, we actually use a variational autoencoder.
We thus use fully-connected layers $\text{MLP}_{\mu},\text{MLP}_{\sigma}$, to embed the graphs into the parameters of a Gaussian distribution.
We also add the standard regularization term into the loss function which imposes a slight penalty towards a standard normal distribution on the learned latent representations.
Then, the vector $\mathbf{z}$ is sampled from the learned Gaussian distribution:
\begin{equation*}
    \boldsymbol{\mu} = \text{MLP}_{\mu}(\mathbf{h}_G), \qquad
    \boldsymbol{\sigma} = \text{MLP}_{\sigma}(\mathbf{h}_G), \qquad
    \mathbf{z} = \boldsymbol{\mu} + \boldsymbol{\sigma} \odot \mathbf{\epsilon}
\end{equation*}
where \(\odot\) denotes element-wise multiplication, and \(\mathbf{\epsilon}\) is a random noise vector sampled from \(\mathcal{N}(0, \mathbf{I})\).

The decoder $\mathcal{D}$ takes the vector $\mathbf{z}$ as input and reconstructs the input graph.
We experiment with a simple MLP decoder that consists of a series of fully connected layers.
Specifically, the decoder is implemented as follows:
\begin{equation*}
    \hat{\mathbf{A}} = \text{MLP}_{\mathcal{D}}(\mathbf{z})
\end{equation*}
Note that $\hat{\mathbf{A}}$ is an $(n_{\text{max}} \times n_{\text{max}})$-dimensional matrix where $n_{\text{max}}$ is a pre-defined maximum graph size.
In fact, the MLP outputs the upper triangular part of this matrix, and from this $\hat{\mathbf{A}}$ is constructed.
Similar to prior work~\cite{de2018molgan}, we found that pre-defining the maximum graph size makes the model significantly faster and also easier to optimize.
Note that $\hat{\mathbf{A}}$ contains continuous values and has a probabilistic interpretation since each element represents the probability that two nodes are connected by an edge.
To produce a discrete object, we use the straight-through Gumbel-Softmax~\cite{jang2017categorical}, \ie we use a sample from a categorical distribution in the forward pass, while the relaxed values are utilized in the backward pass.
The matrix $\hat{\mathbf{A}}$ is then compared with the adjacency matrix of the input graph $\mathbf{A}$ to compute the reconstruction loss.

To compute the reconstruction loss, following previous work, we could employ some graph matching algorithm to compare each input graph against the corresponding reconstructed graph~\cite{simonovsky2018graphvae}.
However, such algorithms are expensive to compute, while we also found empirically that we can achieve similar (or even higher) levels of performance if we just impose a specific ordering on the nodes of each graph and compare the reconstructed adjacency against the adjacency that follows this ordering.
In general, a node ordering that is effective for one type of graphs might not be effective for other types of graphs.
For instance, in the task of molecule generation, certain canonical orderings are more effective than others~\cite{li2018learning}.
Since our dataset consists of graphs that exhibit very different properties from each other, in order to create a general graph generative model, we experimented with the following node orderings: (1) nodes are sorted by degree (from higher to lower); (2) ordering corresponds to BFS/DFS tree's default ordering, which is descending by node degree and rooted at the node with the highest degree; and (3) nodes are sorted by Pagerank scores (from higher to lower).

\subsection{Latent Diffusion Model}
Once the autoencoder is trained, we can use the encoder $\mathcal{E}$ to embed graphs into a low-dimensional
latent space.
Those embeddings are expected to capture both local and global properties of input graphs.
The main advantage of latent diffusion models over standard diffusion models is their efficiency.
The smaller latent space representation makes executing the diffusion process much faster.
This allows the models to be trained on a single or a few GPUs instead of hundreds of GPUs.
Indeed, recent diffusion models for graphs~\cite{vignac2022digress} operate on adjacency tensors (\ie tensors of dimension $n \times n \times d$), and thus are expensive to train, while it is also expensive to generate new graphs.

Latent diffusion models consist of two main components: (1) a noise model; and (2) a denoising
neural network.
The noise model $q$ progressively corrupts the latent representation of the input graph $\mathbf{z}$ to create a sequence of increasingly noisy vectors $(\mathbf{z}_1,\ldots,\mathbf{z}_T)$.
Specifically, the forward process $q$ samples some noise from a Gaussian distribution at each time step $t$, which is added to the representation of the previous time step as follows:
\begin{equation*}
    q(\mathbf{z}_t|\mathbf{z}_{t-1}) = \mathcal{N} \big(\mathbf{z}_t; \sqrt{1- \beta_t} \mathbf{z}_{t-1}, \beta_t \mathbf{I} \big)
\end{equation*}
where $\beta_t$ is a known variance schedule and $0 < \beta_1 < \ldots \beta_T < 1$. 
Note that the forward process $q$ is fixed, and thus, $\mathbf{z}_t$ can be obtained directly from $\mathbf{z}$ during training.
Specifically, we have that:
\begin{equation*}
    q(\mathbf{z}_t|\mathbf{z}) = \mathcal{N} \big(\mathbf{z}_t; \sqrt{\bar{a}_t} \mathbf{z}, (1 - \bar{a}_t) \mathbf{I} \big)
\end{equation*}
where $\bar{a}_t = \prod_{i=1}^t a_i$ and $a_t = 1 - \beta_t$.
Note that if the schedule is set appropriately, $\mathbf{z}_T$ is pure Gaussian noise.
The second component of the latent diffusion model (\ie the denoising neural network) is responsible for predicting the added noise for a given time step $t$.
To train the denoising neural network $\epsilon_\theta$, the following function is minimized:
\begin{equation*}
    \begin{split}
        L_{\text{LDM}} &= \mathbb{E}_{\mathcal{E(G)}, \epsilon \sim \mathcal{N}(0,1),t} \Big[ \big|\big| \epsilon - \epsilon_\theta (\mathbf{z}_t, t) \big|\big|_2^2 \Big]\\
        &= \mathbb{E}_{\mathcal{E(G)}, \epsilon \sim \mathcal{N}(0,1),t} \Big[ \big|\big| \epsilon - \epsilon_\theta (\sqrt{\bar{a}_t} \mathbf{z} + \sqrt{(1-\bar{a}_t)} \epsilon, t) \big|\big|_2^2 \Big]
    \end{split}
\end{equation*}
We implement the denoising neural network $\epsilon_\theta$ as an MLP. 
First, we use sinusoidal positional embeddings to produce a unique representation for each time step $t$.
Then, we feed vector $\mathbf{z}_t$ to the MLP and in each hidden layer of the MLP, we add the positional embedding to the latent representation.

Once the denoising neural network $\epsilon_\theta$ is trained, to generate new data, we sample a vector of pure noise $\mathbf{z}_T$ from a Gaussian distribution, and then use the neural network to gradually denoise it (using the conditional probability it has learned).
The denoised vector can then be transformed into a graph with a single pass through the decoder $\mathcal{D}$.

\subsection{Conditioning}
The ability to condition graph generation on local and global properties of graphs is crucial for the success of our work.
Latent diffusion models are capable of modeling conditional distributions.
Specifically, text, images, or other inputs can be encoded into the latent space and used to condition the model to generate outputs with desired properties.
Given some condition code $\mathbf{c}$, diffusion models are capable of modeling the conditional distribution $p(\mathbf{z}|\mathbf{c})$ by using a conditional neural network $\epsilon_\theta(\mathbf{z}_t, t, \tau_\theta\big(\mathbf{c})\big)$ as follows:
\begin{equation*}
    L_{\text{LDM}} = \mathbb{E}_{\mathcal{E(G)}, \epsilon \sim \mathcal{N}(0,1),t} \Big[ \big|\big| \epsilon - \epsilon_\theta \big(\mathbf{z}_t, t, \tau_\theta(\mathbf{c})\big) \big|\big|_2^2 \Big]
\end{equation*}
where $\epsilon_\theta$ and $\tau_\theta$ are neural network models that are jointly optimized.
The architecture of $\tau_\theta$ depends on the conditioning modality.
As already discussed, the condition code is a vector of properties of the graph.
Thus, we use an MLP as the condition encoder $\tau_\theta$ to compute $\mathbf{y}$ as follows:
\begin{equation*}
    \mathbf{y} = \text{MLP}_c(\mathbf{c})
\end{equation*}
Then, we concatenate vectors $\mathbf{z}$ and $\mathbf{y}$ and feed them into the denoising neural network.
We experimented with other operations (\eg addition, element-wise product), but in preliminary experiments, concatenation led to better results than the other operations.

The condition code consists of $15$ local and global properties of the input graph.
Those properties are listed in Table~\ref{tab:properties} in Appendix~\ref{sec:dataset}.
Those specific $15$ properties were chosen mainly because they cover a broad range of graph properties and they can give insights about the structure of the graph.
Furthermore, the properties can be computed efficiently in polynomial time.

\section{Experimental Evaluation}\label{sec:experiments}
We train and evaluate the NGG model on a dataset that contains $1$M synthetic graphs.
We consider that the evaluation encompasses three distinct scenarios. 
Initially, we assess the model's performance on our original dataset. 
Subsequently, we evaluate its capabilities when trained on graphs containing up to $50$ nodes, test its adaptability to larger graphs with more than $50$ nodes, and examine its generalization performance.
Lastly, we randomly conceal between $1$ to $8$ properties of some of the input graphs (\ie number of nodes, edges, density, etc.), treating them as missing values not provided by the user. 
Subsequently, we retrain the model and assess its performance on the test set, repeating the process of concealing a random number of graph properties.

\paragraph{Dataset.}
We train and evaluate the NGG model on a dataset that contains $1$M synthetic graphs, where each graph contains at most $100$ nodes.
To create the dataset, we use different types of graph generators.
This allows us to construct graphs from different families and thus, cover a wide range of values of the considered properties.
Each synthetic graph belongs to one of the following $17$ families of graphs: (1) paths; (2) cycles; (3) wheels; (4) stars; (5) ladders; (6) lollipops; (7) Erd{\H{o}}s-R{\'e}nyi random graphs; (8) Newman–Watts–Strogatz small-world graphs; (9) Watts–Strogatz small-world graphs, (10) random $d$-degree regular graphs; (11) Barab{\'a}si–Albert graphs; (12) dual Barab{\'a}si–Albert graphs; (13) extended Barab{\'a}si–Albert graphs; (14) graphs generated using the Holme and Kim algorithm; (15) random lobsters; (16) stochastic block model graphs; and (17) random partition graphs.
More details about the constructed dataset are given in Appendix~\ref{sec:dataset}.
The generated graphs in our dataset are devoid of self-loops, isolated nodes, and multigraphs are also excluded.

\paragraph{Experimental setup.}
With regard to the variational autoencoder, the encoder $\mathcal{E}$ consists of $2$ layers, the hidden dimension size is set equal to $64$, while the input graphs are embedded into $32$-dimensional vectors.
We imposed an ordering on the nodes based on the BFS tree's default ordering.
The decoder $\mathcal{D}$ consists of $3$ layers and the hidden dimension size is set to $256$.
The output of $\mathcal{D}$ is a $100^2$-dimensional vector and is transformed into an adjacency matrix of dimension $(100 \times 100)$.
Thus, the model can handle graphs of up to $100$ nodes.
We also initially annotate the nodes of each graph with $10$-dimensional spectral features (from the eigenvectors associated with the $10$ smallest eigenvalues of the normalized Laplacian matrix).
For the latent diffusion model, we set the number of timesteps $T$ equal to $500$, while we use a cosine schedule for adding noise at each time step.
The denoising neural network $\epsilon_\theta$ consists of $3$ layers, the hidden dimension size is equal to $512$, while the size of the output is set to $128$.
To train both models, we use the Adam optimizer and we set the learning rate to $0.001$.
The batch size is set to $256$.
We set the number of epochs of the variational autoencoder to $200$, while the number of epochs of the latent diffusion model to $100$.
Overall, the variational autoencoder consists of $2,640,492$ parameters, while the diffusion model consists of $973,088$ parameters. 
Thus, in total, the NGG model consists of $3,613,580$ parameters. 
The model was pretrained for about 24 hours in a GPU RTX A6000.


\paragraph{Evaluation and baseline.}
Given a collection of graphs $\{ G_1, G_2, \ldots, G_N\}$ that belong to the test set, we use the trained generative model to produce another collection of graphs $\{ \hat{G}_1, \hat{G}_2, \ldots, \hat{G}_N\}$.
To generate each graph of the new collection, we sample pure noise from a Gaussian distribution and use the neural network to gradually denoise it, while the whole denoising process is conditioned on the vector of properties of the corresponding graph of the test set $G_i$.
Once the noise has been removed, the emerging vector is transformed into a graph $\hat{G}_i$ with a single pass through $\mathcal{D}$.
Note, however, that we are not interested in comparing the generated graphs against the corresponding graphs of the test set, but we are interested in finding whether the properties of the generated graphs match the properties of the corresponding graphs of the test set.
Therefore, for each generated graph $\hat{G}_i$, we compute a vector where each component is the value of each one of the $15$ considered properties.
Then, this vector is compared against the vector of properties of $G_i$.
We use two different metrics: ($1$) the mean absolute error (MAE); and ($2$) the symmetric mean absolute percentage error (SMAPE), which we use instead of the mean absolute percentage error (MAPE) because we have values equal to zero or close to zero. 
SMAPE values range from $0\%$ to $100\%$, while lower values indicate better performance, as they represent a smaller percentage difference between the predicted and actual values. 
We calculate the aforementioned metrics by taking into account all $15$ properties and applying z-score normalization to them.

We also use a variant of the variational autoencoder (VGAE) as a baseline where the process is conditioned on vector of properties $\mathbf{c}$ for controlled sampling by concatenating the vector with the graph's latent representation.
Additionally, we conducted further experiments on a subset of the original test set, comparing the performance of the variational autoencoder (VGAE) and our model against a range of LLMs.

\begin{table*}[t]
\begin{center}
\scriptsize
\renewcommand{\arraystretch}{1.2}
\begin{tabular}{l|Hcc|Hcc}
\toprule
\multirow{2}{*}{\textbf{Property}} & \multicolumn{3}{c}{\textbf{VGAE}} & \multicolumn{3}{|c}{\textbf{NGG}} \\
\cmidrule(lr){2-4} \cmidrule(lr){5-7}
 & \textbf{MSE} & \textbf{MAE} & \textbf{SMAPE} & \textbf{MSE} & \textbf{MAE} & \textbf{SMAPE} \\
\midrule
\# nodes & 955.35 & 24.23 & 25.18 & 26.81 & 2.63 & 3.09\\
\# edges & 1,131,294.64 & 701.97 & 66.48 & 9,211.77 & 62.33 & 8.44 \\
Density & 0.15 & 0.32 & 52.13 & 0.003 & 0.04 & 7.23 \\
Min. degree & 486.71 & 13.95 & 64.52 & 362.77 & 11.61 & 49.46 \\
Max. degree & 452.17 & 14.59 & 22.32 & 5.02 & 1.59 & 3.55 \\
Avg. degree & 621.29 & 18.99 & 58.60 & 5.02 & 1.64 & 6.68 \\
Assortativity coefficient & 3.45 & 0.30 & 59.14 & 2.41 & 0.11 & 37.71 \\
\# triangles & 524,208,317.27 & 9,284.64 & 89.57 &  4,073,461.51 & 920.24 & 21.86 \\
Avg. \# triangles formed by an edge  & 215.99 & 7.94 & 59.64 & 311.07 & 8.46 & 48.46 \\
Max. \# triangles formed by an edge  & 745,253.25 & 483.26 & 75.28 & 6,624.62 & 44.17 & 14.94 \\
Avg. local clustering coefficient  & 0.09 & 0.27 & 31.42 & 0.013 & 0.07 & 13.82 \\
Global clustering coefficient & 0.14 & 0.32 & 52.10 & 0.006 & 0.05 & 12.61 \\
Max. $k$-core & 443.89 & 15.07 & 54.51 & 5.33 & 1.66 & 8.61 \\
\# communities & 4.56 & 1.74 & 21.86 & 2.07 & 0.96 & 12.34 \\
Diameter & 75.04 & 1.74 & 21.86 & 54.24 & 2.40 & 15.96 \\
\midrule
All & 301.79 & 1.63 & 56.02  & 297.54 & 1.05 & 21.21\\
\bottomrule
\end{tabular}
\caption{Within distribution performance of the proposed NGG model and the baseline model in terms of the considered properties.}
\label{tab:results_within}
\end{center}
\vskip -0.1in
\end{table*}

\subsection{Within Distribution Performance}
In this experiment, we split the constructed dataset into a training, a validation and a test set.
Then, we train the proposed NGG model and the baseline on the training set.
We choose the models that achieve the lowest loss on the validation set and we evaluate their performance on the test set.
More specifically, the $1$M samples are split into training, validation, and test sets with a $80 : 10 : 10$ split ratio, respectively.
Thus, the training set contains $800,000$ samples, while each of the validation and test sets contain $100,000$ samples.
Each one of the three sets preserves approximately the proportions of each graph type.
Table~\ref{tab:results_within} illustrates the 
MAE and SMAPE achieved by NGG and VGAE for each property individually and for all properties together (last row).
First of all, we observe that the proposed NGG model outperforms VGAE in terms of almost all considered properties.
Secondly, we notice that NGG accurately captures most of the properties, while it struggles with the minimum degree, the number of triangles, and the average and the maximum number of triangles formed by an edge.  

\begin{table*}[t]
\begin{center}
\renewcommand{\arraystretch}{1.2}
\resizebox{\columnwidth}{!}{%
\begin{tabular}{l|Hcc|Hcc|Hcc|Hcc}
\toprule
\multirow{3}{*}{\textbf{Property}} & \multicolumn{6}{c}{\textbf{Out of Distribution}} & \multicolumn{6}{|c}{\textbf{Masked}} \\
\cmidrule(lr){2-7} \cmidrule(lr){8-13}
 & \multicolumn{3}{c|}{\textbf{VGAE}} & \multicolumn{3}{c|}{\textbf{NGG}} & \multicolumn{3}{c|}{\textbf{VGAE}} & \multicolumn{3}{c}{\textbf{NGG}} \\
 & \textbf{MSE} & \textbf{MAE} & \textbf{SMAPE} & \textbf{MSE} & \textbf{MAE} & \textbf{SMAPE} & \textbf{MSE} & \textbf{MAE} & \textbf{SMAPE} & \textbf{MSE} & \textbf{MAE} & \textbf{SMAPE} \\
\midrule
\# nodes & & 15.49 & 18.77 & & 6.69 & 6.67 && 17.89 & 18.13 && 19.28 & 19.45\\
\# edges && 355.79 & 60.13 && 99.09 & 11.32 && 548.98 & 34.84 && 585.83 & 37.24\\
Density && 0.29 & 48.10 && 0.07 & 10.74 && 0.19 & 23.63 && 0.20 & 25.68 \\
Min. degree && 12.59 & 67.12 && 7.95 & 41.49 && 9.70 & 40.09 && 9.31 & 40.07\\
Max degree && 4.46 & 7.95 && 2.94 & 4.64 && 16.51 & 24.22 && 18.14 & 26.52\\
Avg. degree && 14.35 & 51.24 && 3.12 & 9.27 && 13.17 & 28.56 && 14.22 & 30.86\\
Assortativity coefficient && 0.29 & 63.52 && 0.70 & 46.01 && 0.31 & 40.69 && 0.29 & 43.95 \\
\# triangles && 3,562.78 & 76.28 && 1,349.25 & 24.77 && 9,369.57 & 46.48 && 9,941.35 & 49.80 \\
Avg . \# triangles formed by an edge && 7.08 & 50.06 && 8.39 & 50.41 && 11.18 & 46.72 && 11.13 & 48.45\\
Max . \# triangles formed by an edge && 253.39 & 58.72 && 75.69 & 16.12 && 437.86 & 66.55 && 477.43 & 44.61\\
Avg. local clustering coefficient && 0.24 & 23.68 && 0.08 & 14.48 && 0.21 & 25.43 && 0.23 & 27.73 \\
Global clustering coefficient && 0.25 & 42.03 && 0.06 & 13.15 && 0.20 & 25.98 && 0.21 & 27.72\\
Max k-core && 11.61 & 52.76 && 2.28 & 9.99 && 11.79 & 42.96 && 17.99 & 32.05\\
\# communities && 2.65 & 27.47 && 1.02 & 12.32 && 1.29 & 15.82 && 1.36 & 16.24\\
Diameter && 3.53 & 29.99 && 2.55 & 16.62 && 2.35 & 16.44 && 2.31 & 16.89\\
\hline
All && 1.76 & 56.22 && 1.23 & 29.01 && 0.77 & 42.16 && 0.78 & 42.91\\

\bottomrule
\end{tabular}
}
\caption{Performance Comparison of NGG and Baseline Model under Different Conditions: Out-of-Distribution Performance (trained on graphs with up to 50 nodes and evaluated on larger graphs) and Within-Distribution Performance with Masking Applied to some Condition Vector Elements.}
\label{tab:merged_results}
\end{center}
\vskip -0.1in
\end{table*}

\subsection{Out of Distribution Generalization}
In many scenarios, machine learning models fail to generalize to unseen data.
Thus, in this second experiment, we study whether the proposed NGG model and the baseline model can generate graphs whose properties' values are 
different from the ones the models were trained on (i.e. out of distribution).
Specifically, we train the two models on graphs of up to $40$ nodes and also on graphs that consist of more than $60$ nodes.
Then, the models are evaluated on the rest of the graphs (\ie graphs of size from $41$ nodes to $59$ nodes).
The results of this experiment are illustrated in Table~\ref{tab:merged_results} (Left).
We observe that the proposed NGG model still outperforms the VGAE model in this experiment.
However, the VGAE model can generate graphs that better capture the average number of triangles formed by an edge than NGG.
Furthermore, we can also see that both models achieve a bit lower levels of performance compared to their corresponding performance reported in Table~\ref{tab:results_within}.
This is not surprising since the model is trained on graphs that are larger or smaller than the ones on which it is evaluated.

\subsection{Conditioning on Subset of Properties}
Since a practitioner might be interested only in a handful of properties, and not in all $15$ of them, we also investigate whether the proposed model can generate graphs 
providing only a subset of the properties. 
Within each batch, we randomly choose if the properties of each sample would be masked, treating some of the elements of their associated condition vector as missing values as follows. 
For each one of the chosen samples, we randomly choose $i$ elements (without replacement) where $i \in \{1,2,\ldots,8\}$ and replace the values of those elements with a value equal to $-100$.
Note that we also corrupt similarly the samples of the test set at inference time.
We then compare the observed properties of the condition vector against the corresponding properties of the generated samples to compute the three evaluation metrics.
The obtained results are shown in Table~\ref{tab:merged_results} (Right).
We observe that the baseline model outperforms the proposed NGG model, only by a small margin though.
Furthermore, the model achieves significantly worse levels of performance compared to the ones reported in Table~\ref{tab:results_within}.
While the proposed model generally proves effective in the task of graph generation, its performance diminishes when only a subset of the considered properties are available, failing to achieve the same levels of performance as when all properties are present.

\begin{figure*}[t]
    \begin{center}    
    \subfigure[Generated graph that emerged from condition vector $\mathbf{c}_1$]{
        \includegraphics[width=0.45\linewidth]{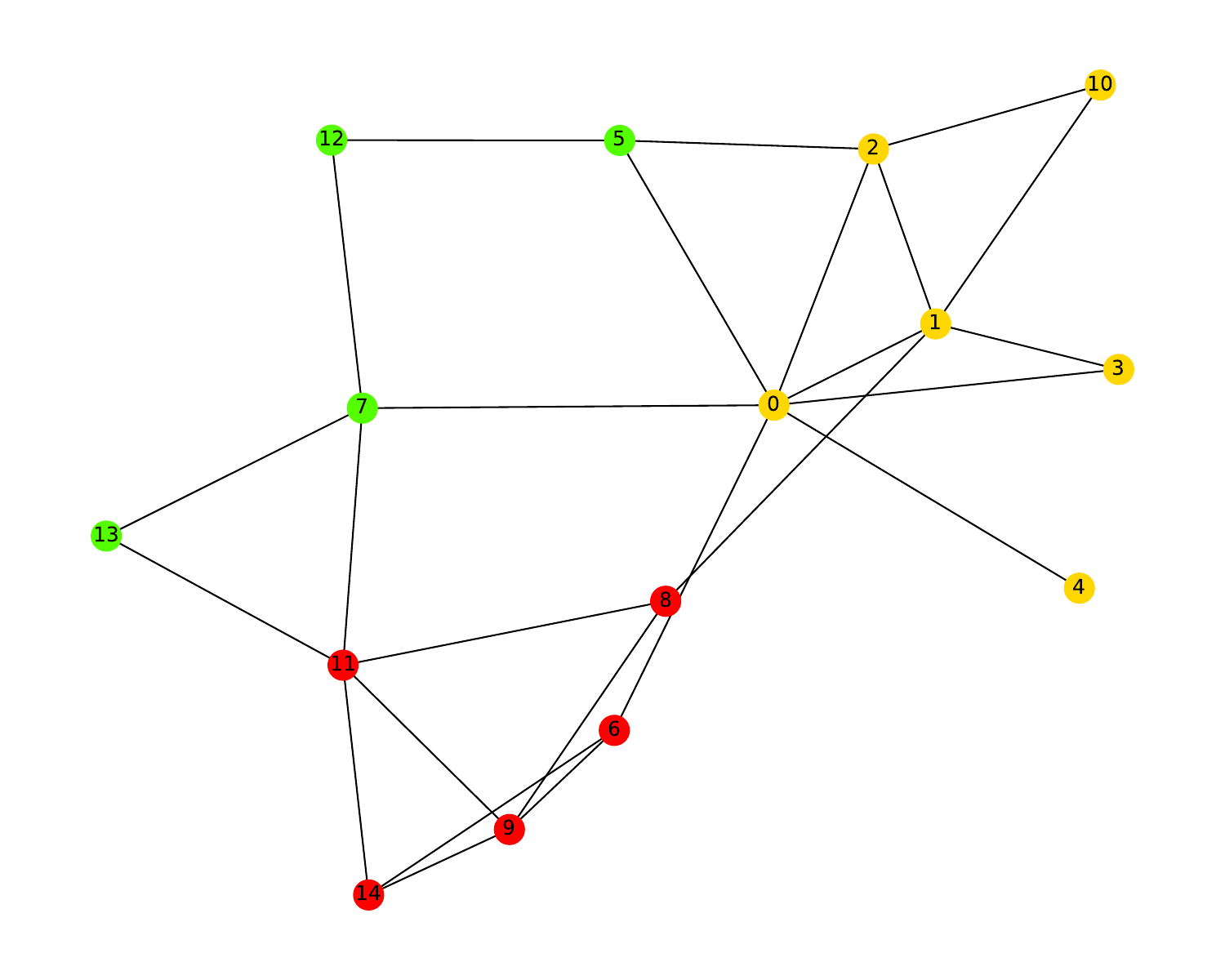}
        \label{fig:example1}
    }
    \hfill
    \subfigure[Generated graph that emerged from condition vector $\mathbf{c}_2$]{
        \includegraphics[width=0.45\linewidth]{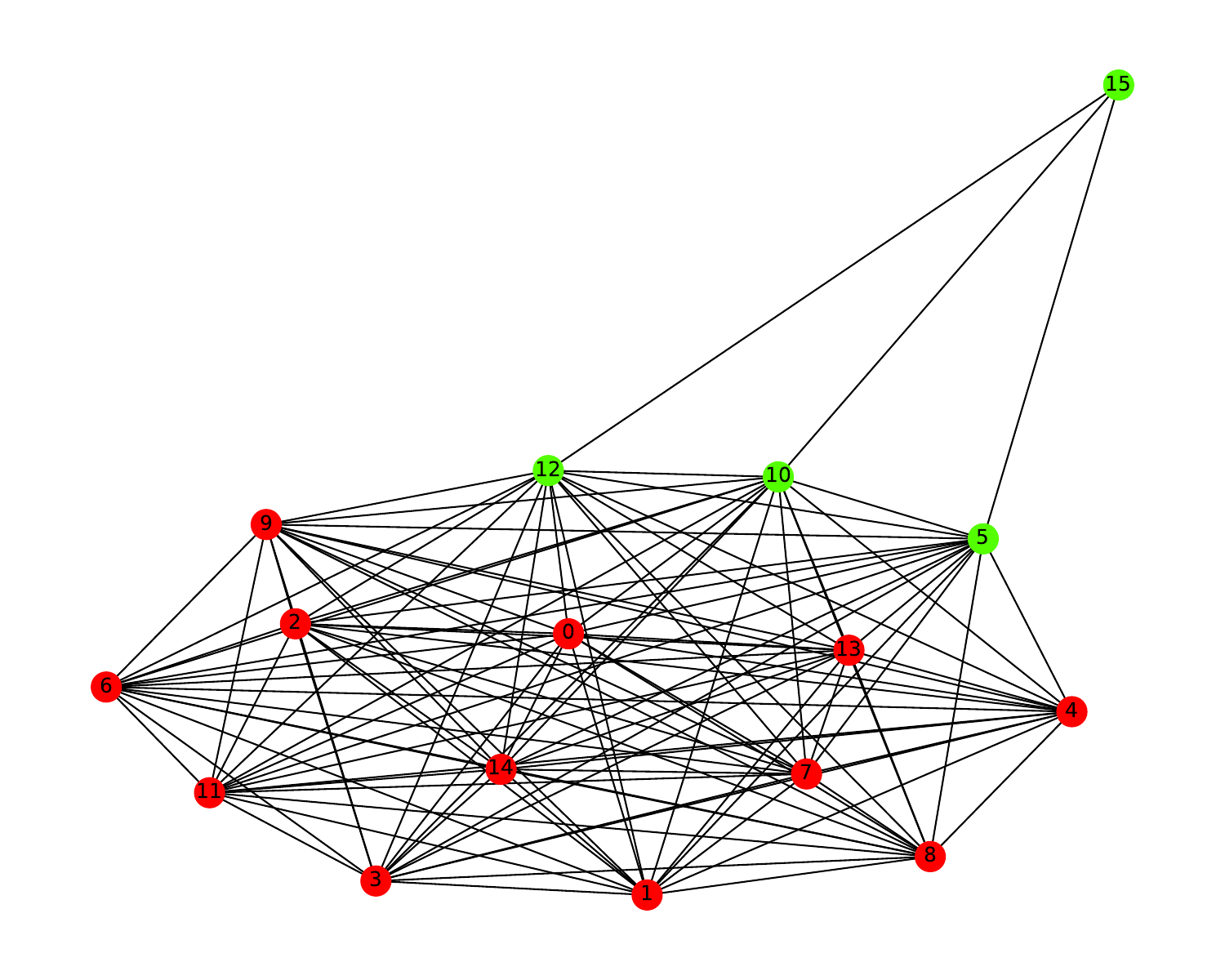}
        \label{fig:example2}
    }    
    \caption{Example of two graphs generated by the proposed NGG model given condition vectors $\mathbf{c}_1$ and $\mathbf{c}_2$.}
    \label{fig:examples}    
    \end{center}
    \vspace{-.2cm}
\end{figure*}

\subsection{NGG vs LLMs}\label{sec:llms}
Recent work has highlighted the expressiveness of large language models (LLMs)~\cite{10.1145/3655103.3655110}, particularly in tasks related to graph generation. In this study, we compared our model and a conditional VGAE against seven LLMs (Llama2-7B, Llama2-13B, Llama-8B, Gemma1.1-2B, Gemma1.1-7B, Mistral-7B, and GPT-3.5-Instruct), evaluating their performance in graph generation, execution time, and GPU usage.
For the experiments, we provided each LLM with the required graph properties and prompted it with generating an edgelist that adhered to these specifications. 
We conducted both \textit{zero-shot }and \textit{one-shot} experiments. In the one-shot scenario, we presented the LLM with the edgelist of a randomly chosen graph with fewer than 20 nodes, along with its 15 properties (same as the ones in the condition vector used above), as an example.
Tables \ref{tab:NGG_LLMs_0shot} and \ref{tab:NGG_LLMs_1shot} present the performance of the LLMs, conditional VGAE, and our model (NGG) on all graph properties. The evaluation was conducted on a subset of 1,000 samples, drawn from the original test set of 100K graphs, due to time and resource constraints. 
The subset was selected such that it reflects the full dataset graph types distribution.
The results in the tables clearly demonstrate that both VGAE and NGG outperform the LLMs across all evaluated criteria (error, execution time and GPU memory usage), with NGG emerging as the top performer.
This illustrated the advantages of the NGG model, achieving superior generative performance but also operating more efficiently, as it is about $10^3$ times faster than LLMs and at the same time consuming fewer resources and being more environmentally friendly. Additionally, it is noteworthy that the one-shot approach underperformed compared to zero-shot for all Llama models, while for Gemma1.1, Mistral, and GPT-3.5, the one-shot method improved the quality of the generated edgelists.

\begin{table}[h!]
    \begin{tabular}{l|c|c|c|c}
        \hline
        Model & MAE & SMAPE (\%) & Execution Time & GPU usage(MB) \\
        \hline
        Llama2-13B & 3.49 & 56.01 & 12h \& 27min & 8,883 \\ 
        Llama2-7B & 2.98 & 54.21 & 14h \& 17min & 4,943 \\
        Llama3-8B & 3.32 & 50.51 & 1d \& 10h & 6,376 \\
        Gemma1.1-2B & 2.60 & 57.59 & 03h \& 38min & 2,455 \\
        Gemma1.1-7B & 2.96 & 57.64 & 02h \& 42min & 6,389 \\
        Mistral-7B & 4.35 & 58.14 & 08h \& 51min & 4,896  \\
        GPT-3.5-Instruct & 7.18 & 52.50 & 04h \& 40min & - \\
        \hline
        VGAE & 1.44 & 48.97 & \textbf{16s} & \textbf{720} \\
        NGG (ours) & \textbf{1.31} & \textbf{43.78} & \textbf{16s} & 800\\     
        \hline
    \end{tabular}
    \caption{Model performance across all graph properties, evaluated on a subset of 1,000 samples from the initial test set (LLMs evaluated with zero-shot prompts).}
    \label{tab:NGG_LLMs_0shot}
\end{table}

\begin{table}[h!]
    \begin{tabular}{l|c|c|c|c}
        \hline
        Model & MAE & SMAPE (\%) & Execution Time & GPU usage(MB) \\
        \hline
        Llama2-13B & 12.35 & 57.28 & 10h \& 46min & 6,722 \\ 
        Llama2-7B & 12.34 & 57.09 & 8h \& 33min & 5,617 \\
        Llama3-8B & 5.74 & 51.16 & 3d \& 20h &  7,818 \\
        Gemma1.1-2B & 2.49 & 59.00 & 1h \& 04min & 2,830 \\
        Gemma1.1-7B & 2.87 & 55.39 & 4h \& 57min & 9,242 \\
        Mistral-7B & 2.39 & 47.55 & 1d \& 20h & 6,873 \\
        GPT-3.5-Instruct & 2.69 & 57.03 & 01h \& 44min & - \\
        \hline
        VGAE & 1.44 & 48.97 & \textbf{16s} & \textbf{720} \\
        NGG (ours) & \textbf{1.31} & \textbf{43.78} & \textbf{16s} & 800\\     
        \hline
    \end{tabular}
    \caption{Model performance across all graph properties, evaluated on a subset of 1,000 samples from the initial test set (LLMs evaluated with one-shot prompts).}
    \label{tab:NGG_LLMs_1shot}
\end{table}

\subsection{Examples of Generated Graphs}\label{sec:examples}
We will next give two examples of graphs generated by the proposed NGG model.
To generate the two graphs, we used the model that was trained on $80\%$ of the entire dataset.
For ease of presentation, we decided to generate two graphs that both consist of $15$ nodes.
However, the second graph is much more dense than the first graph.
Specifically, we utilized the following two condition vectors:
\begin{equation*}
    \begin{split}
    \mathbf{c}_1 &= 
    \begin{psmallmatrix}
        15 & 34 & 0.32 & 2 & 8 & 4.5 & -0.046 & 17 & 1.5 & 8 & 0.4 & 0.35 & 4 & 4 & 4 \\
    \end{psmallmatrix}^\top\\
    \mathbf{c}_2 &= 
    \begin{psmallmatrix}
        15 & 94 & 0.89 & 10 & 14 & 12.5 & -0.149 & 329 & 10.5 & 80 & 0.9 & 0.9 & 10 & 2 & 2
    \end{psmallmatrix}^\top
    \end{split}
\end{equation*}
where each dimension corresponds to one of the 15 properties respectively. 
The above two vectors gave rise to the two graphs that are shown in Figure~\ref{fig:examples}.
The corresponding communities for each generated graph are also illustrated (node color denotes the community to which the node belongs).
We observe that one generated graph indeed consists of $15$ nodes, while the other consists of $16$ nodes.
Furthermore, the graph shown in Figure~\ref{fig:example1} is indeed sparse, while the one shown in Figure~\ref{fig:example2} is dense.
We also computed the values of the considered properties for both graphs and these are given below:
\begin{equation*}
    \begin{split}
    \hat{\mathbf{c}}_1 &= 
    \begin{psmallmatrix}
        15 & 25 & 0.23 & 1 & 7 & 3.3 & -0.380 & 8 & 0.96 & 3 & 0.43 & 0.32 & 2 & 3 & 4
    \end{psmallmatrix}^\top\\
    \hat{\mathbf{c}}_2 &= 
    \begin{psmallmatrix}
        16 & 108 & 0.9 & 3 & 15 & 13.5 & -0.148 & 458 & 12.7 & 93 & 0.97 & 0.97 & 14 & 2 & 2 
    \end{psmallmatrix}^\top
    \end{split}
\end{equation*}
We can see that for most properties, their actual values are close to the ones that were utilized for generating the two graphs.
To summarize, those results provide a qualitative validation of the proposed NGG model.
We also visually inspected several other graphs and it turns out that the proposed model can generate graphs that approximately exhibit the desired properties.

\subsection{Uniqueness of Generated Graphs}
We also investigated whether given a vector of properties, the proposed NGG model can generate non-isomorphic graphs.
Specifically, we randomly chose $50$ condition codes, and for each code, we used the model to produce $100$ graphs.
We then tested every pair of these graphs for isomorphism ($9900/2$ pairs in total for each condition code).
We found that all generated graphs are unique, \ie there was no pair of isomorphic graphs.
Therefore, the NGG model seems to also produce diverse sets of graphs.

\subsection{Limitation of baselines}

In our work, we chose not to conduct experiments comparing our model with the DiGress~\cite{vignac2022digress} or EDGE~\cite{10.5555/3618408.3618589} or other state-of-the-art generative models. The primary reason for this decision is 
that these architectures are not compatible with the conditioning/prompting mechanism we propose involving multiple fuatures of a graph. 

If we would want to compare directly we should adapt these models to our specific conditioning mechanism and synthetic dataset. Implementing such adaptations would not only be technically demanding but would also represent a new contribution, beyond the scope of our current study. We leave it as potential future work though that could explore these adaptations in detail, potentially leading to novel contributions in the application of DiGress and EDGE to similar datasets.

\section{Conclusion}\label{sec:conclusion}
We introduced the Neural Graph Generator (NGG), a conditional latent diffusion model, for efficient and accurate graph generation. NGG represents a significant new paradigm, adeptly generating graphs conditioned on a vector of multiple  specific, user-defined properties, a task that has long challenged traditional models. 
We demonstrated the ability of NGG to generalize to graphs beyond those in the training set, and the ability to handle missing values in the condition vector.
We also highlighted the importance of a dedicated graph generator, particularly in scenarios where LLMs struggle to capture complex structural representations. 
Our model not only outperforms LLMs in graph generation tasks but also does so more efficiently, consuming fewer resources.
Nevertheless, GNN models, while powerful, struggle to capture specific properties like triangles or cycles, thus affecting the performance of our model. 
Future work will focus on exploring the NGG's extensions/adaptations to various real-world scenarios (\ie power/telecom networks, molecules/protein graphs), and scaling it to larger graphs.

\bibliographystyle{apalike}
\bibliography{main}

\appendix
\section{Dataset Details}\label{sec:dataset}
The distribution of the samples per family of graphs is illustrated in Table~\ref{tab:dataset_distribution}. 

\begin{table*}[h!]
\vskip 0.15in
\begin{center}
\footnotesize
\renewcommand{\arraystretch}{1.4}
    \begin{tabular}{p{8cm}|c|c} 
    \toprule
        \textbf{Type of graph} & \textbf{Number of samples} & \textbf{Proportion} \\
        \midrule
        Barabási–Albert random graph & 250,136 & 25.01\% \\
        Watts–Strogatz small-world graph & 204,280 & 20.43\% \\
        stochastic block model graph & 125,468 & 12.55\% \\
        Erdős-Rényi graph & 122,568 & 12.26\% \\
        dual Barabási–Albert random graph & 122,568 & 12.26\% \\
        extended Barabási–Albert model graph & 122,568 & 12.26\% \\
        Newman-Watts-Strogatz small-world graph & 122,567 & 12.26\% \\
        Holme and Kim algorithm for growing graphs with powerlaw degree distribution and approximate average clustering & 122,567 & 12.26\% \\
        Random Lobster graph & 81,713 & 8.17\% \\
        random partition graph & 81,713 & 8.17\% \\ 
        Random $d$-degree regular graph & 7,000 & 0.7\% \\ 
        Lollipop graph & 4,145 & 0.41\% \\    
        Path graph & 91 & 0.01\% \\
        Cycle graph & 91 & 0.01\% \\
        Star graph & 91 & 0.01\% \\
        Wheel graph & 91 & 0.01\% \\
        Ladder graph & 46 & 0.0046\% \\         
    \bottomrule
\end{tabular}
\caption{Distribution of samples per graph type}
\label{tab:dataset_distribution}
\end{center}
\vspace{.8cm}
\end{table*}

\newpage
The $15$ considered graph properties and their description are listed in Table~\ref{tab:properties}.

\begin{table*}[t]
\begin{center}
\renewcommand{\arraystretch}{1.0}
\resizebox{\columnwidth}{!}{%
    \begin{tabular}{p{4cm}|c|p{11cm}} 
    \toprule
    \textbf{Property} & \textbf{Range} & \textbf{Description} \\
    \midrule
    \# nodes & [2,100] & Count of individual vertices or entities in a graph. \\
    \midrule
    \# edges & [1,4950] & Count of connections or relationships between nodes in the graph.\\
    \midrule
    Density & [0.01,1] & A measure of how interconnected a graph is, calculated as the number of actual edges over the total possible edges.\\
    \midrule
    Min. degree & [1,99] &  Minimum degree is the smallest degree among all nodes.\\
    \midrule
    Max. degree & [1,99] &  Maximum degree is the largest degree among all nodes.\\
    \midrule
    Avg. degree & [1,99] & Mean degree across all nodes.\\
    \midrule
    Assortativity coefficient & [-1,1] & A measure of the tendency of nodes to connect with others of similar degree. Positive values suggest a preference for connections between nodes of similar degrees.\\
    \midrule
    \# triangles & [0, 161700] & Number of sets of three nodes that form a closed loop (triangle) in the network.\\
    \midrule
    Avg. \# triangles formed by an edge & [0, 98] & Average number of triangles each edge participates in.\\
    \midrule
    Max. \# triangles formed by an edge & [0, 4851] & Highest number of triangles a single edge is part of.\\
    \midrule
    Avg. local clustering coefficient & [0, 1] & A measure of the extent to which nodes in a neighborhood tend to form clusters or cliques.\\
    \midrule
    Global clustering coefficient & [0, 1] & A measure of the overall tendency of nodes to form clusters in the entire graph, reflecting the global structure of the network.\\
    \midrule
    Max. $k$-core & [0, 99] & Largest subgraph where each node has at least $k$ neighbors within the subgraph.\\
    \midrule
    \# communities & [0, 50] & Number of clusters of nodes (communities) in the graph.\\
    \midrule
    Diameter & [0, 99] & Maximum distance between any pair of nodes in the graph.\\
    \bottomrule
\end{tabular}
}
\caption{Description of the $15$ considered graph properties.}
\label{tab:properties}
\end{center}
\vskip -0.1in
\end{table*}

\section{LLM prompting examples}

In tables \ref{tab:NGG_LLMs_0shot} and \ref{tab:NGG_LLMs_1shot}, it is clear that we employed different output lengths in terms of tokens, varying both by the specific LLM used and by the learning approach (zero-shot or one-shot). This variation was essential to optimize performance within the constraints of each model and task.
The smaller LLMs require less GPU memory, allowing us to generate more tokens during inference.
For models like ChatGPT, we determined that the output length is governed by the formula: \texttt{output length = 4k - input length}.
Consequently, the maximum allowable output length in each scenario is constrained by the available resources or API limitations.

\begin{figure*}[h!]
    \begin{center}    
    \subfigure[Example of zero-shot prompting]{
        \includegraphics[width=0.9\linewidth]{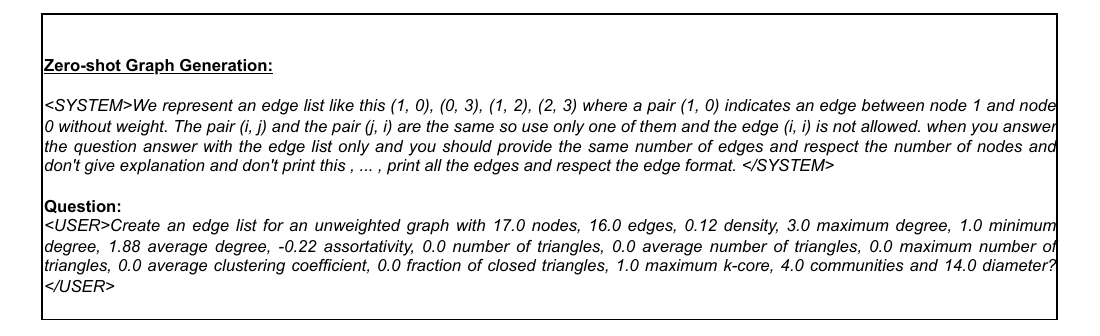}
        \label{fig:0shot_example}
    }
    \hfill
    \subfigure[Example of one-shot prompting]{
        \includegraphics[width=0.9\linewidth]{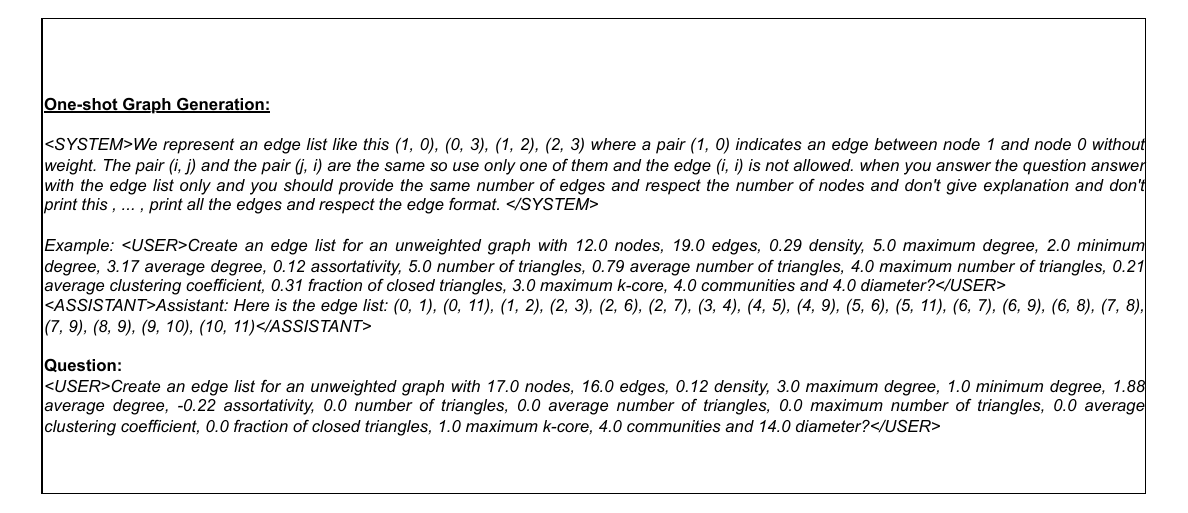}
        \label{fig:1shot_example}
    }    
    \caption{Prompting examples for LLM graph generation}
    \label{fig:prompt_examples}    
    \end{center}
\end{figure*}

Figures \ref{fig:0shot_example} and \ref{fig:1shot_example} provide examples of generalized prompting for zero-shot and one-shot learning, respectively.


\section{Ablation Study}\label{sec:ablation}


\subsection{Different methods for the construction of adjacency}
Table~\ref{tab:ordering} illustrates the impact of different node ordering techniques on the construction of the adjacency matrix and the resulting performance of the model. The results indicate that initiating the BFS algorithm from the node with the highest degree is the most effective approach. This method consistently leads to better performance, highlighting its prominence in structuring the adjacency matrix.
Interestingly, when the adjacency matrix is constructed starting from the node with the highest PageRank score, there is a risk of generating a few invalid graphs during inference (about $0.002\%$ invalid graphs). This suggests that while PageRank-based ordering can be useful, it may introduce complications that undermine the model's reliability.
Additionally, ordering the nodes based on degree without the BFS algorithm also yields good results, but it slightly underperforms compared to the BFS-based approach. This observation underscores the advantage of using BFS with degree-based root selection in optimizing graph representation and model accuracy.

\begin{table*}[t]
\label{tab:ordering}
\begin{center}
\renewcommand{\arraystretch}{1.0}
\resizebox{\columnwidth}{!}{%
    \begin{tabular}{p{4cm}|c|c|p{4cm}|} 
    \toprule
    \textbf{Method} & \textbf{MAE} & \textbf{SMAPE (\%)} & \textbf{Percentage of Non-Valid Graphs (\%)}\\
    \midrule
    BFS \& Degree & \textbf{1.05} & \textbf{21.21} & \textbf{0}\\
    Degree only & 1.78 & 21.64 & \textbf{0}\\
    Pagerank & 1.69 & 22.18 & 0.002 \\
    \midrule
\end{tabular}
}
\caption{Performance of different ordering methods for the construction of adjacency matrix}
\end{center}
\vskip -0.1in
\end{table*}

\newpage
\subsection{Importance of Graph Properties}

\begin{figure*}[h!]
    \centering
    \centerline{\includegraphics[width=1.\linewidth]{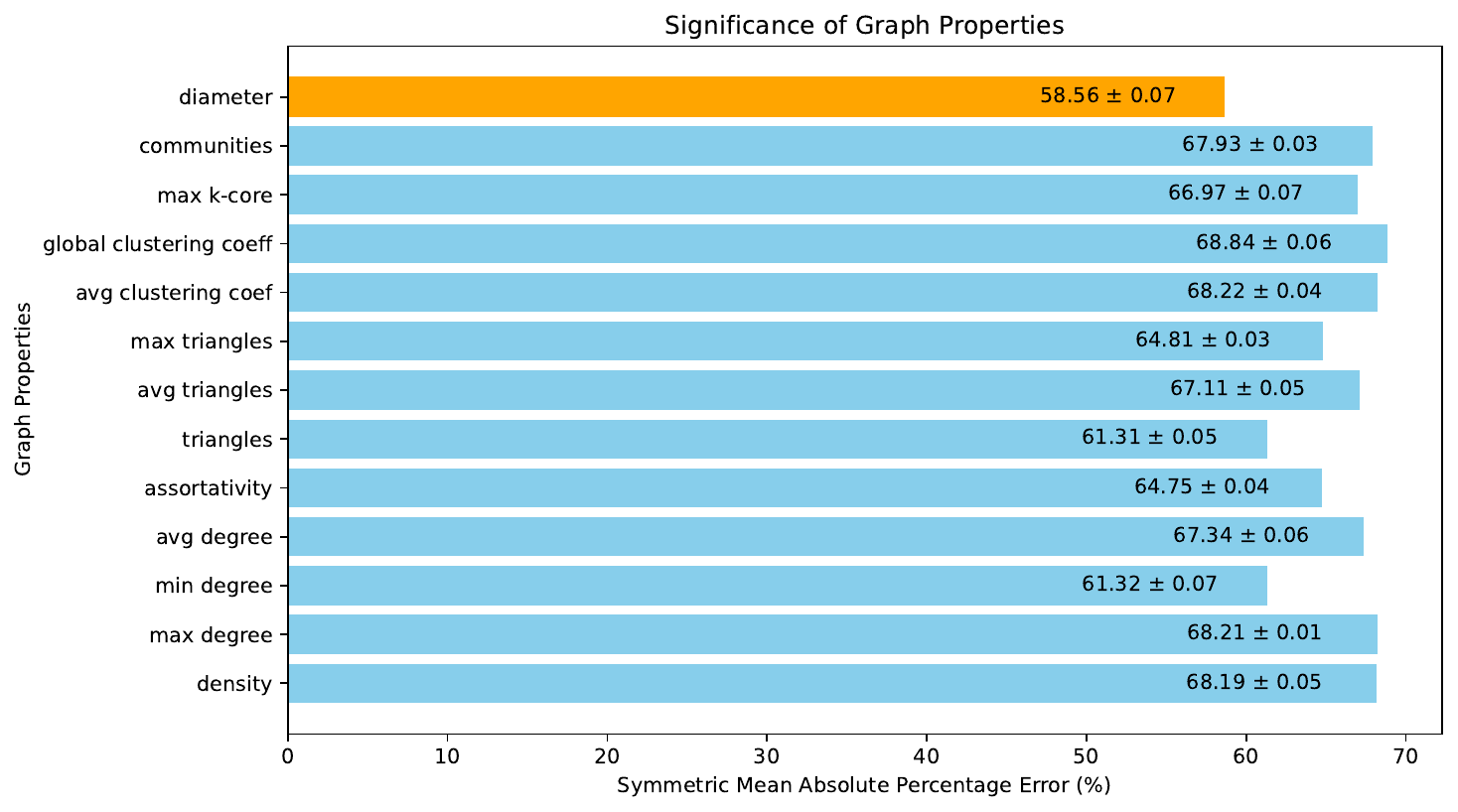}}
    \caption{Importance of graph properties, given the number of nodes, edges, and one of the remaining 13 properties as input.}
    \label{fig:properties}
    \vspace{-.2cm}
\end{figure*}

Given that we already know the number of nodes and edges, we investigated the significance of individual graph properties when only one of the remaining 13 properties is known (including the number of nodes and edges). In these experiments, we provided the model with the number of nodes, the number of edges, and one additional property, effectively creating a triplet of properties while masking the other 12.
We conducted these experiments across the entire test set, repeating each trial three times to minimize the impact of random variation on the results. The mean SMAPE metric from these three runs is reported as the outcome of this ablation study.
In Figure \ref{fig:properties}, the property highlighted in orange is the one that contributes most significantly to the model's performance. This analysis helps identify which properties, when combined with the number of nodes and edges, have the greatest impact on the model's performance.



\end{document}